\documentclass[letterpaper]{article} 
\usepackage{aaai24}  
\usepackage{times}  
\usepackage{helvet}  
\usepackage{courier}  
\usepackage[hyphens]{url}  
\usepackage{graphicx} 
\usepackage{natbib}  
\usepackage{caption} 
\usepackage{algorithm}
\usepackage{algorithmic}

\usepackage{amsmath}
\usepackage{bm}
\usepackage{subcaption}
\newtheorem{proposition}{Proposition}

\usepackage{multirow}

\usepackage{newfloat}
\usepackage{listings}
\DeclareCaptionStyle{ruled}{labelfont=normalfont,labelsep=colon,strut=off} 
\floatstyle{ruled}
\newfloat{listing}{tb}{lst}{}
\floatname{listing}{Listing}
\title{E2E-AT: A Unified Framework for Tackling Uncertainty \\ in Task-aware End-to-end Learning}
\author {
    Wangkun Xu\textsuperscript{\rm 1},
    Jianhong Wang\textsuperscript{\rm 2},
    Fei Teng\textsuperscript{\rm 1}
}
\affiliations {
    \textsuperscript{\rm 1}Department of EEE, Imperial College London, UK \\
    \textsuperscript{\rm 2}Center for AI Fundamentals, The University of Manchester, UK \\
    wangkun.xu18@imperial.ac.uk, jianhong.wang@manchester.ac.uk, fei.teng@imperial.ac.uk
}

\usepackage{bibentry}

\begin{document}

\maketitle

\begin{abstract}
Successful machine learning involves a complete pipeline of data, model, and downstream applications. Instead of treating them separately, there has been a prominent increase of attention within the constrained optimization (CO) and machine learning (ML) communities towards combining prediction and optimization models. The so-called end-to-end (E2E) learning captures the task-based objective for which they will be used for decision making. Although a large variety of E2E algorithms have been presented, it has not been fully investigated how to systematically address uncertainties involved in such models. Most of the existing work considers the uncertainties of ML in the input space and improves robustness through adversarial training. We extend this idea to E2E learning and prove that there is a robustness certification procedure by solving augmented integer programming. Furthermore, we show that neglecting the uncertainty of COs during training causes a new trigger for generalization errors. To include all these components, we propose a unified framework that covers the uncertainties emerging in both the input feature space of the ML models and the COs. The framework is described as a robust optimization problem and is practically solved via end-to-end adversarial training (E2E-AT). Finally, the performance of E2E-AT is evaluated by a real-world end-to-end power system operation problem, including load forecasting and sequential scheduling tasks\footnote{Accepted by AAAI-24. Our code is available at \url{https://github.com/xuwkk/E2E-AT}.}.
\end{abstract}

\section{Introduction}

ML-based prediction algorithms are widely used in real-world applications, such as load forecasting in power system operation, demand forecasting in retailing, and inventory stock forecasting in commerce. Although training such models is straightforwardly a supervised learning problem, the training criteria may not be aligned with the ultimate goal of the downstream decision-making tasks. For instance, the forecast load is further used for generator dispatch in the power system, and the demand forecasting can be used to guide manufacturing. That is, the system designer first trains an ML forecaster using standard statistic training loss, e.g. the mean squared error (MSE) or cross entropy (CE), and then applies the forecast as input to the decision making. In fact, task-aware cost, which is modeled as a constrained optimization (CO) problem, can benefit from training a more economical forecast model. Therefore, task-aware end-to-end learning is proposed by combining ML and COs. However, the uncertainties involved in E2E learning have not been fully understood. This paper aims to bridge the gap from the perspective of robust optimization and adversarial robustness.

In the literature, E2E learning is also named as (smart) predict-then-optimize \cite{elmachtoub2022smart}, integrated learning and optimization \cite{sadana2023survey}, as well as decision-focused learning \cite{wilder2019melding}. In the paradigm of contextual optimization, it appears as early as in \cite{bengio1997using} and has recently shown a surge of interest. The E2E model is integrated to minimize the task-aware cost subject to task constraints. As a result, the CO is first solved in the forward pass to obtain the optimal decision. In the backward pass, the Jacobian from the optimal decision to the forecast value needs to be calculated. Among many approaches, this paper adopts the implicit differentiation method, which is more efficient than the unrolling approach \cite{domke2012generic} and more accurate than the surrogate loss function approach \cite{elmachtoub2022smart}. Practically, OptNet \cite{amos2017optnet} applies the implicit function theorem \cite{krantz2002implicit} to denote the Jacobian after formulating the Karush-Kuhn-Ticher (KKT) condition of parametric quadratic programming (QP), which is further extended to disciplined convex programming in CvxpyLayers \cite{agrawal2019differentiable}. 

E2E learning has been applied to many critical industrial activities, such as better management of the power system \cite{stratigakos2022prescriptive, vohra2023end}, constraint satisfaction in control \cite{chen2021enforcing}, and routing behavior of the transportation network \cite{liu2023end}. Therefore, it is essential to understand the vulnerability of this learning framework and to study the associated robust enhancement. In detail, the E2E model contains three parts: (a). a parametric forecast model that maps the contextual information (e.g. the input feature) to the interest of forecast; (b). CO models that take the forecast as input and return decisions; and (c). a task-aware loss function that encodes the ultimate goal of decision making. Meanwhile, the parameters of COs and task-aware loss can be classified into predictable and unpredictable parameters. While the uncertainty of predictable parameters can be modeled by the forecast model, the uncertainty of unpredictable parameters are overlooked in literature. Conventionally, the uncertainty of the data is only defined by the input of the forecast model, the worst case of which can be found by adversarial attack and treated by adversarial training \cite{madry2017towards}. In the E2E model, the definition of data is augmented by the unpredictable parameters of COs\footnote{We merge the unpredictable parameter of task-aware cost as part of CO.}. Our main contribution is to treat multiple sources of uncertainty uniformly as a robust optimization problem, which can be practically solved by end-to-end adversarial training (E2E-AT). Finally, our method can be viewed as a natural interconnection between adversarial training, the implicit layer, and E2E learning.


\section{Preliminaries}

\subsection{Adversarial Training and Certified Robustness}

In supervised learning, a parametric model $\bm{f}(\bm{x};\bm{\theta})$ can be defined to map from input $\bm{x}$ to label $\bm{y}$ for $(\bm{x},\bm{y})\in\mathcal{D}$ by minimizing the following empirical loss:
\begin{equation}\label{eq:supervised_learning}
    \operatorname{min}_{\bm{\theta}} \sum_{i\in\mathcal{D}}\mathcal{L}(\bm{f}(\bm{x}^i; \bm{\theta}), \bm{y}^i)
\end{equation}

In this paper, we use $(\bm{x},\bm{y})\in\mathcal{D}$ and $i\in\mathcal{D}$ interchangeably to denote a sample of the dataset. 

It has been well studied that the deep neural network is prone to small perturbations on its input. In security-critical applications, robustness has become an emerging factor. Adversarial training has been shown to improve the robustness of NN, which transforms \eqref{eq:supervised_learning} into a robust optimization \cite{madry2017towards}:
\begin{equation}\label{eq:adversarial_training}
    \operatorname{min}_{\bm{\theta}} \sum_{i\in\mathcal{D}} \operatorname{max}_{\bm{\delta}_i\in\Delta}\mathcal{L}(\bm{f}(\bm{x}^i + \bm{\delta}^i; \bm{\theta}), \bm{y}^i)
\end{equation}
where $\Delta = \{\bm{\delta}:\|\bm{\delta}\|_\infty \leq \epsilon\}$ is the attack budget set for some small $\epsilon > 0$.

Robust optimization \eqref{eq:adversarial_training} is practically solved by iterative approaches in which the adversarial perturbation is first solved by inner minimization through gradient ascent. To keep the attack within $\Delta$, projected gradient descent (PGD) \cite{madry2017towards} is adopted:
\begin{equation}\label{eq:pgd}
    \bm{\delta}^i_{t+1} = \mathcal{P}_{\Delta}\left(\bm{\delta}_t^i+\gamma \cdot \operatorname{sign}\left(\nabla_{\bm{\delta}} \mathcal{L}\left(\bm{f}\left(\bm{x}^i+\bm{\delta}^i_t\right), \bm{y}^i\right)\right)\right)
\end{equation}
where $\mathcal{P}_{\Delta}$ is the projector on $\Delta$ and $\gamma$ is the step size. 

Although adversarial training is effective in improving robustness, it may give the wrong sign of security as inner maximization is inexactly solved. A certification approach can be made on the feedforward neural network with ReLU activations, e.g., a piecewise linear neural network\footnote{Convolution layer and other piecewise linear activations, such as leaky ReLU, are also piecewise linear.}. Using the big-M method, the neural network with $d$ layers can be represented by the following set \cite{tjeng2017evaluating}
\begin{equation}\label{eq:nn_milp}
    \mathcal{C}_{\text{nn}}(\bm{x};\bm{\theta}) = \left\{\bm{y}: \begin{array}{l}
        \bm{z}_1 = \bm{x}, \bm{z}_{i+1} \geq \bm{W}_i \bm{z}_i + \bm{b}_i, \\
        \bm{z}_{i+1} \geq \bm{0}, \bm{u}_i \cdot \bm{v}_i \geq  \bm{z}_{i+1} \\
        \bm{W}_i \bm{z}_i + \bm{b}_i \geq \bm{z}_{i+1} + (\bm{1}-\bm{v}_i) \bm{l}_i, \\
        \bm{v}_i \in \{0,1\}^{|\bm{v}_i|},\quad i = 1,\cdots,d-2 \\
        \bm{y} = \bm{W}_{d-1}\bm{z}_{d-1} + \bm{b}_{d-1}
    \end{array}\right\}
\end{equation}
where $\bm{\theta}_i = (\bm{W}_i, \bm{b}_i)$ is the weights of the $i$-th layer. $\bm{u}_i$ and $\bm{l}_i$ are the upper and lower bounds of the output of $i$-th layer, which can be efficiently estimated by interval bound propagation (IBP) \cite{gowal2018effectiveness}. $\bm{v}_i$ is an integer vector that controls the activation of ReLUs. 

Based on \eqref{eq:nn_milp}, the inner maximization of \eqref{eq:adversarial_training} becomes:
\begin{equation}\label{eq:attack_exact}
    \operatorname{max}_{\bm{\delta}^i}\mathcal{L}(\hat{\bm{y}}^i, \bm{y}^i), \text{ subject to } \bm{\delta}^i\in\Delta, \hat{\bm{y}}^i\in\mathcal{C}_{\text{nn}}(\bm{x}^i+\bm{\delta}^i;\bm{\theta})
\end{equation}

As $\mathcal{C}_{\text{nn}}(\bm{x};\bm{\theta})$ is defined by mixed integer linear constraints, \eqref{eq:attack_exact} can be solved exactly for certain types of loss function. When cross-entropy loss is used, we can choose to maximize the output of each logit, and if all the logits are not greater than the ground-truth logit within the attack budget, the NN is said to be certified robust at the candidate sample. For a regression task, \eqref{eq:attack_exact} can be formulated as mixed-integer linear programming (MILP) to maximize or minimize the forecast value \cite{xu2023availability}.

\subsection{End-to-End Machine Learning}

End-to-end learning takes the fusion of prediction (ML) and decision making (CO), which aims to find the map from the data to the optimal decision such that the learned ML model can optimally reflect the CO via training. Meanwhile, the E2E learning distinguishes itself from separate supervised learning followed by the CO in which the prediction divergence and the task-aware cost can have a large mismatch \cite{sadana2023survey}. 

Referring to a recent review \cite{kotary2021end}, the CO can be modelled as 
\begin{equation}\label{eq:co}
    \bm{z}^\star = \operatorname{argmin}_{\bm{z}} \ell(\bm{z};\bm{y}) \text{  subject to  } \bm{z} \in \mathcal{C}(\bm{z};\bm{y})
\end{equation}
where $\ell(\cdot)$ is the objective function, $\bm{z}$ is the decision variable, $\bm{y}$ is the parameter, and $\mathcal{C}$ is the constraint set.

Let $\hat{\bm{y}}$ be the output of $\bm{f}(\bm{x};\bm{\theta})$ and let $\hat{\bm{z}}^\star$ be the minimizer of \eqref{eq:co} parameterized by $\hat{\bm{y}}$. There are two options to learn the E2E model guided by \eqref{eq:co}. In the supervised learning setting, a suitable loss function $\mathcal{M}$ can be chosen to minimize the difference to ground-truth decision, e.g. $\mathcal{M}(\hat{\bm{z}}^\star, \bm{z}^\star )$ or $\mathcal{M}(\ell(\hat{\bm{z}}^\star;\hat{\bm{y}}), \ell(\bm{z}^\star;\bm{y}))$ \cite{kong2022end} . Alternatively, the regret function \cite{elmachtoub2022smart} can be implemented: 
\begin{equation}\label{eq:regret}
    \operatorname{regret}=\ell(\hat{\bm{z}}^\star; \hat{\bm{y}})-\ell(\bm{z}^\star; \bm{y})
\end{equation}

The ground truth decision $\bm{z}^\star$ in \eqref{eq:regret} is not necessary to compute, as $\ell(\bm{z}^\star; \bm{y})$ is a constant \cite{wilder2019melding}. Therefore, we argue that the regret function \eqref{eq:regret} is more intrinsic as the training loss has the same format as the CO objective \eqref{eq:co}.

\section{A Heuristic Example}

Before we move into a detailed formulation of the robustness of E2E learning, we first highlight a misleading formulation, which violates the intention of E2E learning and introduces model uncertainties between training and inference. 

Consider two E2E learning problems based on the supervised setting:
\begin{equation}\label{eq:form_one}
    \begin{array}{rl}
        \operatorname{argmin}_{\bm{\vartheta}, \hat{\bm{z}}} &  M(\hat{\bm{z}}) := \sum_{i \in \mathcal{D}} \mathcal{M}(\ell(\hat{\bm{z}}^{i}; \hat{\bm{y}}^i), \ell(\bm{z}^{i\star}; \bm{y}^i)) \\
        \text{subject to} & \hat{\bm{z}}^i \in \mathcal{C}(\bm{z}^i;\hat{\bm{y}}^i), \quad i\in\mathcal{D} \\
    \end{array}
\end{equation}
whose minimum point is denoted as $(\bm{\theta}^\star_1,\hat{\bm{z}}^\star_1)$.
And,
\begin{equation}\label{eq:form_two}
    \begin{array}{rl}
        \operatorname{argmin}_{\bm{\vartheta}, \hat{\bm{z}}} & M(\hat{\bm{z}}) :=  \sum_{i \in \mathcal{D}} \mathcal{M}(\ell(\hat{\bm{z}}^i; \hat{\bm{y}}^i), \ell(\bm{z}^{i\star}; \bm{y}^i)) \\
        \text{subject to} & \hat{\bm{z}}^i \in \operatorname{argmin}_{\bm{z}^i}\{\ell(\bm{z}^i;\hat{\bm{y}}^i): \bm{z}^i\in\mathcal{C}(\bm{z}^i;\hat{\bm{y}}^i)\}, \\
        & \qquad \qquad \qquad \qquad \qquad \qquad  i\in\mathcal{D}
    \end{array}
\end{equation}
with minimum point $(\bm{\theta}^\star_2,\hat{\bm{z}}^\star_2)$.

E2E learning problem \eqref{eq:form_one} takes the sample-wise constraints of \eqref{eq:co} as its constraints while \eqref{eq:form_two} is a bilevel optimization problem whose upper level minimizes the difference to the ground-truth objective $\ell(\bm{z}^{i\star}; \bm{y}^i)$ and each lower level solves the sample-wise decision-making problem in parallel. 

In the \textbf{inference stage}, we first predict $\hat{\bm{y}} = f(\bm{x};\bm{\theta}^\star)$ and then solve the optimization problem \eqref{eq:co} parameterized by $\hat{\bm{y}}$. The optimal decision variables are denoted as $\hat{\bm{z}}^\star_r(\bm{\theta}^\star_1)$ and $\hat{\bm{z}}^\star_r(\bm{\theta}^\star_2)$, respectively. 

\begin{proposition}\label{theorem:misleading}
    Given two formulations \eqref{eq:form_one} and \eqref{eq:form_two} of E2E learning, $M(\hat{\bm{z}}_1^\star) \leq M(\hat{\bm{z}}_2^\star) = M(\hat{\bm{z}}_r^\star(\bm{\theta}_2^\star)) \leq M(\hat{\bm{z}}_r^\star(\bm{\theta}_1^\star)) $.
\end{proposition}

The proof can be found in the appendix.

It shows that formulation \eqref{eq:form_one} can result in different decisions at the training and inference stages. Therefore, Proposition \ref{theorem:misleading} demonstrates that a misformulation of CO can possibly lead to misled decision making at inference. Broadly speaking, the ignorance of the optimization model contributes to a new source of uncertainties, causing generalization error, in addition to the well-studied uncertainties in the dataset (e.g. out-of-distribution, adversarial attack, etc.). This motivates us to take into account the uncertainties in both the input space and the CO formulation.

\section{Sequential E2E Learning as Multi-Level Optimization}

Proposition \ref{theorem:misleading} implies formulating E2E learning as a multilevel optimization problem by respecting the sequence of downstream tasks. Given $\bm{\theta}$, sequential decision makings can be denoted as lower-level problems after the prediction. 

Formally, at the inference stage, the cost of one-time decision making on $(\bm{x},\bm{y})\in\mathcal{D}$ can be written as:
\begin{equation}\label{eq:spo_inference}
    \begin{array}{rl}
        \operatorname{min} & \mathcal{L}(\hat{\bm{z}}_1,\cdots, \hat{\bm{z}}_m ; \bm{y}, \bm{\phi}_0) \\
        \text{subject to} & \hat{\bm{z}}_i\in \operatorname{argmin}_{\bm{z}_i} \{\ell_i(\bm{z}_i; \bm{y}, \bm{\phi}_i): \\
        & \qquad \bm{z}_i\in\mathcal{C}_i(\bm{z}_i; \bm{y}, \hat{\bm{z}}_{i-1}, \bm{\phi}_i)\}, \; i=2,\cdots,m \\
        & \hat{\bm{z}}_1\in \operatorname{argmin}_{\bm{z}_1} \{\ell_1(\bm{z}_1;\bm{y},\bm{\phi}_1): \\
        & \qquad \bm{z}_1\in\mathcal{C}_1(\bm{z}_1 ; \bm{y}, \hat{\bm{y}},\bm{\phi}_1)\} \\
        & \hat{\bm{y}} = \bm{f}(\bm{x};\bm{\theta})
    \end{array}
\end{equation}

Inference \eqref{eq:spo_inference} contains $m$ downstream tasks with objective $\ell_i(\cdot)$s, constraint set $\mathcal{C}_i(\cdot)$s, and \textbf{unpredictable} parameter $\bm{\phi}_i$s. It can be compactly denoted as
\begin{equation}\label{eq:spo_inference_compact}
    \begin{array}{ll}
        \operatorname{min}_{\hat{\bm{z}}} & \mathcal{L}(\hat{\bm{z}}; \bm{\theta}, \bm{x}, \bm{y}, \bm{\phi}) \\
        \text{subject to} & \hat{\bm{z}} \in \mathcal{C}_{\text{E2E}}(\bm{z}; \bm{\theta}, \bm{x}, \bm{y}, \bm{\phi})
    \end{array}
\end{equation}
where $\hat{\bm{z}} = [\hat{\bm{z}}_1, \cdots, \hat{\bm{z}}_m]$ and $\bm{\phi}=[\bm{\phi}_0,\cdots,\bm{\phi}_m]$. $\mathcal{C}_{\text{E2E}}(\cdot)$ represents the constraint set representing all constraints of \eqref{eq:spo_inference}. Note that we consider the COs as parametric functions that can be uniquely modeled by $(\bm{\theta}, \bm{x}, \bm{y}, \bm{\phi})$. 

To train $\bm{f}(\cdot;\bm{\theta})$, the empirical training loss can be minimized:
\begin{equation}\label{eq:spo_learning}
    \begin{array}{ll}
        \textrm{min}_{\bm{\theta}} & \sum_{i\in\mathcal{D}} \mathcal{L}(\hat{\bm{z}}^i; \bm{\theta}, \bm{x}^i, \bm{y}^i, \bm{\phi}) \\
        \text{subject to} & \hat{\bm{z}}^i \in \mathcal{C}_{\text{E2E}}(\bm{z}^i; \bm{\theta}, \bm{x}^i, \bm{y}^i, \bm{\phi}), \quad i\in\mathcal{D}
    \end{array}
\end{equation}

Since this paper does not focus on solving optimizations, we restrict downstream optimizations to quadratic programming (QP). Furthermore, QP has been widely implemented in many industrial applications and is mostly discussed in the E2E learning literature \cite{kotary2021end}. Meanwhile, since QP is convex and if the Slater condition holds, the Karush–Kuhn–Tucker (KKT) condition is sufficient for optimality \cite{boyd2004convex}. Therefore, optimizations at the lower level can be replaced by the corresponding KKT conditions, known as the mathematical program with equilibrium constraints (MPEC) \cite{luo1996mathematical}. Therefore, if a linear parametric model is considered, it is possible to solve \eqref{eq:spo_learning} exactly using optimization software.

In addition to the linear model, stochastic gradient descent (SGD) needs to be applied on the mini-batches of $\mathcal{D}$ to train the NN model. SGD requires 1) a forward pass in which the optimizations are solved and 2) a backward pass to update the NN parameters. Denote the equality part of KKT condition of the $i$-th optimization as
\begin{equation}\label{eq:kkt_layer}
    \bm{g}(\bm{z}^\star_i; \hat{\bm{z}}_{i-1}, \bm{y}, \bm{\phi}_i) = \bm{0}
\end{equation}
which can be viewed as differentiable layer (OptNet) \cite{amos2017optnet} by the implicit function theorem \cite{krantz2002implicit}:
\begin{equation}\label{eq:implicit}
    \frac{\partial \bm{z}_i^\star}{\partial \hat{\bm{z}}_{i-1}} = - \left(\frac{\partial \bm{g}(\bm{z}^\star_i;\hat{\bm{z}}_{i-1}, \bm{y}, \bm{\phi}_i)}{\partial \bm{z}_i}\right)^{-1}\frac{\partial \bm{g}(\bm{z}^\star_i;\hat{\bm{z}}_{i-1}, \bm{y}, \bm{\phi}_i)}{\partial \hat{\bm{z}}_{i-1}}
\end{equation}

Note that in \eqref{eq:kkt_layer} and \eqref{eq:implicit}, the optimal dual variables $(\bm{\lambda}^\star,\bm{\nu}^\star)$ in the KKT conditions are omitted for simplicity. As long as the Jacobian matrix is not singular, the gradient of the output $\bm{z}_{i}^\star$ to the input $\hat{\bm{z}}_{i-1}$ exists, allowing backpropagation through the differentiable layers.

\section{Unified Robustness Framework}

\begin{figure}
    \centering
    \includegraphics[width=0.96\linewidth]{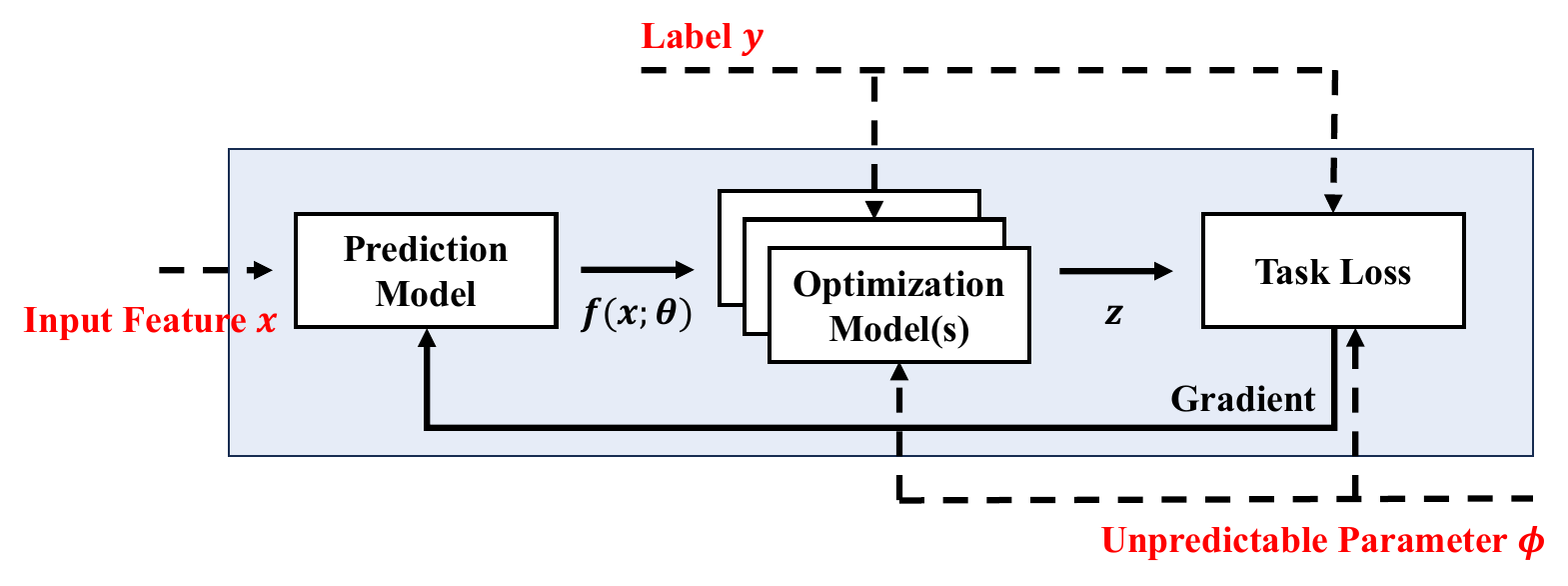}
    \caption{An illustration of E2E learning. We consider the uncertainties in the input data $(\bm{x},\bm{y})\in\mathcal{D}$ and the uncertainties in the COs, specifically, the unpredictable parameter $\bm{\phi}$.}
    \label{fig:e2e_flowchart}
\end{figure}

When treating the E2E framework as an integrated model, the data source includes both conventionally defined data samples $(\bm{x},\bm{y})\in\mathcal{D}$ and the unpredictable parameter $\bm{\phi}$ of COs. Small input uncertainties have been shown to cause a significant performance drop, and it is reasonable to draw a similar conclusion for the unpredictable parameter. In fact, Proposition \ref{theorem:misleading} shows that any mismatches between the COs used for training and inference should be explicitly considered. Although COs can take an infinite number of formulations, without loss of generality, we restrict the uncertainties of COs in the parameters of objective and constraints. We argue that the unpredictable parameter used during training may not be the same as that used for real-time decision-making. For example, in power system operation, the production costs of the generators can vary over time, and the resistance and susceptance of transmission lines can be altered both intentionally and unintentionally. These parameters are usually not known to the system operator in advance or at least not fully aware when training the forecast model. See Fig. \ref{fig:e2e_flowchart} for an illustration. 

\subsection{Formulation}

Consider the uncertainty of the input $\bm{x} + \bm{\delta}_x\in\mathcal{X}$ and the unpredictable parameter $\bm{\phi} + \bm{\delta}_\phi \in \Phi$. Denote $\bm{\psi} = (\bm{x}, \bm{\phi}) \in \Psi := \mathcal{X} \times \Phi$. The worst scenario, which maximizes the task-aware objective, can be formulated as
\begin{equation}\label{eq:spo_adversarial_attack}
    \begin{array}{ll}
         \operatorname{max}_{\bm{\psi} \in \Psi} & \mathcal{L}(\hat{\bm{z}};\bm{\theta},\bm{\psi},\bm{y})   \\
         \text{subject to } &  \hat{\bm{z}} \in \mathcal{C}_{\text{E2E}}(\bm{z}; \bm{\theta}, \bm{\psi},\bm{y})
    \end{array}
\end{equation}
in which $\bm{\theta}$ is fixed.

Consequently, a robust optimization can be formulated. A unified E2E adversarial training (E2E-AT) considering both input and CO uncertainties becomes
\begin{equation}\label{eq:robust_formulation}
    \begin{array}{rl}
         \operatorname{min}_{\bm{\theta}} & \sum_{i\in\mathcal{D}} \operatorname{max}_{\bm{\psi}^i \in \Psi^i} \mathcal{L}(\{\hat{\bm{z}}^i;\bm{\theta},\bm{\psi}^i,\bm{y}^i)   \\
         \text{subject to } &  \hat{\bm{z}}^i \in \mathcal{C}_{\text{E2E}}(\bm{z}^i; \bm{\theta}, \bm{\psi}^i,\bm{y}^i), \quad i\in\mathcal{D}
    \end{array}
\end{equation}
where the constrains are subject to both minimization and maximization. Adversarial training can be adopted to solve \eqref{eq:robust_formulation}. Similarly to the implicit function theorem \eqref{eq:implicit}, the gradient of the constraint exists, regardless of the minimization or maximization. Therefore, Danskin's theorem can be used by first solving the inner maximization through gradient ascent (with $\bm{\theta}$ fixed) and then for the outer gradient descent (with $\bm{\psi}$ fixed). Although using the Danskin theorem requires one to exactly solve the inner maximization, it can give a descent direction for suboptimal $\bm{\psi}$, e.g. solved by PGD \eqref{eq:pgd} and is applicable to various adversarial training algorithms \cite{madry2017towards, zhang2019theoretically, dong2020adversarial}.

\subsection{Certified Robustness}

Although PGD \eqref{eq:pgd} is effective for E2E-AT, it cannot verify the robustness, as it only finds the local maximum \cite{xiao2018training}. In particular, a robustness certification on \eqref{eq:spo_inference_compact} verifies if an adversarial example exists within the budget $\Delta$ such that the task-aware objective is altered by a certain amount. The key is to find the exact adversarial attack in \eqref{eq:spo_adversarial_attack}. We show that for specific type of objective and COs (e.g. affine-parametric QPs), optimal solution to \eqref{eq:spo_adversarial_attack} can be solved exactly, which extends the certified robustness in piecewise linear neural network \eqref{eq:nn_milp}. 

\begin{proposition}\label{theorem:qp_kkt}
    The affine-parametric QP:
    \begin{equation}\label{eq:qp}
    \begin{array}{rll}
        \bm{z}_{i+1} := & \arg\min_{\bm{z}} & \frac{1}{2}\bm{z}^T\bm{Q}\bm{z} + \bm{q}^T\bm{z} \\
        & \mathrm{subject \ to} &  \bm{A}\bm{z} + \bm{G}\bm{z}_i \leq \bm{b} \\
        & & \bm{C}\bm{z} + \bm{H}\bm{z}_i = \bm{d}
    \end{array}
    \end{equation}
    can be equivalently written as the set of mixed integer linear constraints:
    \begin{subequations}
    \begin{equation*}\label{eq:kkt_stationary}
        \bm{Q}\bm{z}_{i+1} + \bm{q} + \bm{A}^T\bm{\lambda}_{i+1} + \bm{C}^T\bm{\nu}_{i+1} = 0 
    \end{equation*}
    \begin{equation*}\label{eq:kkt_equality}
        \bm{C}\bm{z}_{i+1} + \bm{H}\bm{z}_i - \bm{d} = \bm{0}
    \end{equation*}
    \begin{equation*}
        \bm{\lambda}_{i+1} \geq \bm{0}
    \end{equation*}
    \begin{equation*}
        \bm{A}\bm{z}_{i+1} + \bm{G}\bm{z}_i - \bm{b} \leq \bm{0}
    \end{equation*}
    \begin{equation*}
        \bm{\lambda}_{i+1}\leq\bm{\varphi}\bm{M}, \; \bm{A}\bm{z}_{i+1} + \bm{G}\bm{z}_i - \bm{b} \geq (\bm{\varphi} - \bm{1}) \bm{M}, \; \bm{\varphi}\in\{0,1\}^{|\bm{\varphi}|}
    \end{equation*}
\end{subequations}
where $\bm{Q}, \bm{q}, \bm{A}, \bm{G}, \bm{b}, \bm{C}, \bm{H}, \bm{d}$ are the parameters with proper dimensions. $\bm{\varphi}$ is binary vector and $\bm{M}$ is a large positive number. The equalities and inequalities are element-wise.

\end{proposition}

The proof can be found in the appendix.

Due to the complexity of integer programming, we restrict the original settings in \cite{amos2017optnet} by assuming linearity in uncertain terms for certified robustness. For example, when the uncertainty in the input feature is considered, the parameter $\bm{z}_i$ that represents the optimal decision of the previous task is decoupled from the variable $\bm{z}$ and is affine so that the reformulation is linear. When the uncertainty of CO is considered, we assume that the uncertain unpredictable parameter is decoupled from the variable as well. We note that this setting follows the disciplined parametrized programming (DPP) \cite{agrawal2019differentiable}.

Since both affine-parametric QP \eqref{eq:qp} and neural network \eqref{eq:nn_milp} can be represented by mixed-integer linear constraints, maximizing an affine function subject to these constraints becomes mixed-integer linear programming (MILP), which can be effectively solved and used to certify the worst possible cost.

\subsection{Discussion on the Robustness}

Previously, the uncertainty involved in COs has been considered in many E2E learning algorithms. From the perspective of contextual optimization, E2E learning applies ML to predict the uncertain parameter \cite{sadana2023survey}. In such setting, probabilistic forecast and stochastic program can be implemented \cite{donti2017task}. However, we view the entire E2E learning as an integrated model such that the uncertainty of the intermediate variable can be merged within the E2E training objective. Alternatively, we separate the parameters of COs into two parts. The first part (predictable parameter) is forecasted by the ML while the second part is unpredictable. Indirectly, the uncertainty of the predictable parameter is tackled by adversarial training on the input feature, while the uncertainty of unpredictable parameter needs to be tackled as well.

We note that the physical meaning of the uncertainty of the unpredictable parameter is different from that of the adversarial attack in the input feature space. Although the ultimate goal is to improve the robustness of the ML, the adversarial attack assumes that there exists a malicious party that can find the worst attack. However, the uncertainty of the unpredictable parameter always exists regardless of the malicious party. Inspired by the previous work \cite{donti2021adversarially, agarwal2022employing}, we can alternatively view the E2E-AT on the unpredictable parameter as the following optimization problem:
\begin{equation}\label{eq:spo_learning_opt}
    \begin{array}{rl}
        \textrm{min}_{\bm{\theta}} & \quad \alpha \sum_{i\in\mathcal{D}} \mathcal{L}(\bar{\bm{z}}^i; \bm{\theta}, \bm{x}^i, \bm{y}^i, \bar{\bm{\phi}}) \\
        & + (1- \alpha) \sum_{i\in\mathcal{D}} \mathbf{E}_{(\hat{\bm{z}}^i,\bm{\phi}^i)}[\mathcal{L}(\hat{\bm{z}}^i; \bm{\theta}, \bm{x}^i, \bm{y}^i, \bm{\phi}^i)] \\
        \text{subject to} & \bar{\bm{z}}^i \in \mathcal{C}_{\text{E2E}}(\bm{z}^i; \bm{\theta}, \bm{x}^i, \bm{y}^i, \bar{\bm{\phi}}), \;  i\in\mathcal{D}\\
        & \hat{\bm{z}}^i \in \mathcal{C}_{\text{E2E}}(\bm{z}^i; \bm{\theta}, \bm{x}^i, \bm{y}^i, \bm{\phi}^i), \;  \bm{\phi}^i\in{\Phi}^i, i\in\mathcal{D}
    \end{array}
\end{equation}
where $\bar{\bm{z}}^i$ is the decision variable from the COs parameterized by the nominal unpredictable parameter $\bar{\bm{\phi}}$.

It can be argued that the NN forecaster is trained by considering the expected task-aware cost due to uncertainties. The learning objective \eqref{eq:spo_learning_opt} takes the nominal unpredictable parameter (denoted as $\bar{\bm{\phi}}$) and the expected uncertainties over $\bm{\Phi}^i$ into account, which are balanced by the hyperparameter $\alpha$. As shown in \eqref{eq:spo_learning_opt}, this stochastic program can be solved by sampling $\bm{\phi}^i\in\bm{\Phi}^i$ during training. In addition, if the uncertainty set of the unpredictable parameter is independent of the sample, $\bm{\Phi}$ is not subject to index $i$. 

\subsection{Final Training Objective}

In E2E-AT, \eqref{eq:spo_learning_opt} is solved by robust optimization by finding the maximum over $\bm{\psi}^i \in \Psi^i$, as in \eqref{eq:robust_formulation}. We also take the input uncertainty into account:
\begin{equation}\label{eq:robust_formulation_balance}
    \begin{array}{rl}
         \operatorname{min}_{\bm{\theta}} & 
         \quad \alpha \cdot \sum_{i\in\mathcal{D}} \mathcal{L}(\bar{\bm{z}}^i;\bm{\theta}, \bar{\bm{\psi}}^i, \bm{y}^i) \\
         & + (1-\alpha) \cdot \sum_{i\in\mathcal{D}} \operatorname{max}_{\bm{\psi}^i \in \Psi^i} \mathcal{L}(\hat{\bm{z}}^i;\bm{\theta},\bm{\psi}^i,\bm{y}^i)   \\
         \text{subject to} & \bar{\bm{z}}^i \in \mathcal{C}_{\text{E2E}}(\bm{z}^i; \bm{\theta}, \bar{\bm{\psi}} , \bm{y}^i) \\
        & \hat{\bm{z}}^i \in \mathcal{C}_{\text{E2E}}(\bm{z}^i; \bm{\theta}, \bm{\psi}^i, \bm{y}^i), \quad i\in\mathcal{D}
    \end{array}
\end{equation}

The new adversarial training objective provides an upper bound on the expectation part of \eqref{eq:spo_learning_opt}. Meanwhile, similar to adversarial training on image tasks \cite{zhang2019theoretically}, $\alpha$ can be used to balance the clean and adversarial accuracies. When $\alpha\rightarrow 1$, \eqref{eq:robust_formulation_balance} becomes the original E2E learning and when $\alpha\rightarrow 0$, it becomes pure adversarial training. In addition, clean and adversarial training losses may not directly reflect clean and robust accuracy for image tasks, causing an unbalanced training objective. In E2E-AT, the two objectives are defined by the task, which is the exact metric during decision making.

Previously in \cite{donti2021adversarially, agarwal2022employing}, the authors reformulate a bilevel optimization problem into robust optimization and use the implicit function theorem for constraint satisfaction. Connecting \eqref{eq:spo_learning_opt} to \eqref{eq:robust_formulation_balance}, we extend \cite{donti2021adversarially} to training ML models. It can be seen that for each mini-batch in E2E-AT, a similar robust optimization is solved as \cite{donti2021adversarially}. The implicit function theorem is also used to learn the task-aware objective while satisfying the constraints.

\subsection{`Free' E2E Adversarial Training}

In E2E-AT, the number of forward and backward passes is equal to $\texttt{no\_batch} \times (\texttt{no\_pgd} + 1) \times \texttt{no\_epoch}$ (assuming that all minibatches have the same size). Adversarial training is computationally ineffective due to intensive backpropagation (controlled by the complexity of the neural network). The computational burden is even higher in E2E-AT as in each forward pass, the COs need to be solved (controlled by the complexity of the COs). To save training time, we adopt the gradient reuse strategy. In the \textit{adversarial training for free} \cite{shafahi2019adversarial}, the attack vector and the model parameter are repeatedly updated in the same mini-batch for $\texttt{no\_pgd}$ times. Then, $\texttt{epoch\_no}$ is divided by $\texttt{no\_pgd}$ to maintain the total number of model updates unchanged. This results in $\texttt{no\_batch} \times \texttt{no\_epoch}$ numbers of forward and backward passes, which is the same as the clean E2E training.

\section{Experiment}

In the experiment, the NN is trained to forecast the load in the power system. The robustness of various E2E-AT settings is explored. Detailed experiment settings and results can be found in the Appendix.

\subsection{Power System Operations}

A practical power system operation problem, named as network constrained economic dispatch (NCED), is considered, which has been widely used in the US and can be formulated as two-stage QP or LP \cite{conejo2018power}. In stage one (also known as dispatch), the set points of the generator are determined based on the \textbf{forecast} load. The goal of the first stage is to minimize the generator cost while meeting the physical constraints of the grid. When the generator has been dispatched, we consider a realization on the actual load by solving the second stage problem (also known as redispatch), in which any mismatches on the load and generation from stage one, as well as the violation of the physical constraints of the grid, will be penalized by extra cost:
\begin{subequations}
    \begin{equation}
        \bm{P}_g^\star = \text{Dispatch}(\bm{f}(\bm{x};\bm{\theta}), \bm{b})
    \end{equation}
    \begin{equation}
        \bm{P}_{ls}^\star, \bm{P}_{gs}^\star = \text{Redispatch}(\bm{y}, \bm{P}_g^\star, \bm{b})
    \end{equation}
\end{subequations}
where $\bm{P}_g^\star$ is the optimal generator dispatch, $\bm{P}_{ls}^\star$ is the load shedding and $\bm{P}_{gs}^\star$ is the power storage. $\bm{b}$ is the susceptance of the transmission line (when the resistance is close to zero, the susceptance is reciprocal to the reactance). The task-aware objective is defined as
\begin{equation}\label{eq:training_objective}
    \mathcal{L}(\bm{\theta}) = \bm{c}_g^T\bm{P}_{g} + \bm{c}_{ls}^T\bm{P}_{ls} + \bm{c}_{gs}^T\bm{P}_{gs}
\end{equation}
where $\bm{c}_g$, $\bm{c}_{ls}$, and $\bm{c}_{gs}$ are the coefficients such that $\bm{c}_{ls}\gg\bm{c}_{gs}\gg\bm{c}_g$. That is, the load shedding is more costly.

\subsection{Training Settings}

\begin{table*}[t]
    \centering
    \caption{Performances of the E2E-AT.}
    \scalebox{0.95}{
    \begin{tabular}{c|c|c|c|c|c|c|c|c|c|c|c|c}\hline
       \multicolumn{3}{c|}{\textbf{Training Method}} & \textbf{Clean} & \multicolumn{3}{c|}{\textbf{Input Attack}, $\epsilon_x$} & \multicolumn{3}{c|}{\textbf{CO Attack}, $\epsilon_\phi$} & \multicolumn{3}{c}{\textbf{Integrated Attack}, ($\epsilon_x, \epsilon_\phi$)}   \\\hline\hline
       $\epsilon_x$ & $\epsilon_\phi$ & $\alpha$ & N/A & 0.01 & 0.02 & 0.03 & 0.05 & 0.10 & 0.15 & (0.01,0.05) & (0.02,0.10) & (0.03,0.15) \\\hline\hline

       \multicolumn{13}{c}{\textbf{NAT: Natural Training with Task Loss}} \\\hline\hline
       0 & 0 & 0 & 201.6 & 653.6 & 1188.5 & 1792.1 & 754.6 & 2044.3 & 3468.4 & 1153.9 & 3090.8 & 5214.4 \\\hline\hline
       
        \multicolumn{13}{c}{\textbf{AT-MSE: Adversarial Training with MSE Loss}} \\\hline\hline
       0.02 & N/A & N/A & 396.5 & 676.6 & 978.3 & 1278.4 & 1223.8 & 2576.7 & 3923.6 & 1475.4 & 3120.1 & 4778.3  \\\hline
       0.03 & N/A & N/A & 437.0 & 653.6 & 886.3 & 1125.5 & 1208.8 & 2517.2 & 3881.6 & 1410.5 & 2956.5 & 4624.0 \\\hline\hline
       
       \multicolumn{13}{c}{\textbf{AT-INPUT: Adversarial Training with Task Loss on the Input Uncertainty}} \\\hline\hline
       \multirow{2}{*}{0.02} & \multirow{2}{*}{0} & 1.0 & 219.1 & 445.2 & 745.0 & 1106.1 & 672.4 & 1929.0 & 3334.02 & 934.4 & 2706.3 & 4854.5\\\cline{3-13}
                                                 &  & 0.5 & 203.1 & 438.9 & 790.1 & 1246.7 & 660.7 & 1932.7 & 3349.6 & 914.8 & 2763.9 & 4930.5\\\hline
       \multirow{2}{*}{0.03} & \multirow{2}{*}{0} & 1.0 & 239.4  & 448.6 & 705.0 & 1022.1 & 638.7 & 1787.5 & 3220.6 & 855.5 & 2445.9 & 4507.0\\\cline{3-13}
                                                 &  & 0.5 & 227.6  & 556.8 & 976.6 & 1435.3 & 783.7 & 2074.9 & 3503.1 & 1106.8 & 2948.5 &  4999.8  \\\hline\hline
                                                 
       \multicolumn{13}{c}{\textbf{AT-PARA: Adversarial Training with Task Loss on the CO Uncertainty}} \\\hline\hline
       \multirow{2}{*}{0} & \multirow{2}{*}{0.05} & 1.0 & 212.3 & 398.84 & 726.8 & 1197.4 & 214.9 & 590.1  & 1931.3 & 399.1 & 939.1 & 2620.0 \\\cline{3-13}
                                             &  & 0.5 & 206.7  & 419.3  & 798.2 & 1299.9 & 215.6  & 741.2 & 2120.4 & 419.8 & 1114.8 & 3072.6 \\\hline
       \multirow{2}{*}{0} & \multirow{2}{*}{0.15} & 1.0 & 214.0 & 488.7 & 900.1 & 1434.5 & 214.0 & 221.0 & 453.2 & 493.0 & 925.8 & 1502.0 \\\cline{3-13}
                                             &  & 0.5 & 216.5 & 423.5 & 804.6 & 1309.8 & 216.5 & 223.1 & 465.6 & 429.6 & 817.8 & 1324.6 \\\hline\hline

       \multicolumn{13}{c}{\textbf{AT-BOTH: Adversarial Training with Task Loss on the Integrated Uncertainties}} \\\hline\hline
       \multirow{2}{*}{0.02} & \multirow{2}{*}{0.05} & 1.0 & 228.2 & 399.5 & 655.7 & 992.0 &  254.2 & 823.5 & 2156.7 & 400.5  & 1128.7 & 2956.2 \\\cline{3-13}
                                                 &  & 0.5 & 212.1 & 490.1 & 879.9 & 1336.5 & 289.4 & 1257.9 & 2711.8 & 516.0 & 1793.9 & 3903.8 \\\hline
       \multirow{2}{*}{0.03} & \multirow{2}{*}{0.15} & 1.0 & 274.7 & 551.0 & 856.5 & 1226.3 & 274.7 & 279.0 & 495.7 & 550.6 & 883.4 & 1274.0 \\\cline{3-13}
                                                 &  & 0.5 & 239.2 & 429.1 & 666.3 & 977.6 & 239.2 & 250.8 & 508.3 & 429.0 & 667.6 & 1053.7 \\\hline
    \end{tabular}
    }
    \label{tab:adverarial}
\end{table*}

We use an open source load forecasting dataset from the Texas Backbone Power System \cite{lu2023synthetic} on a modified IEEE bus-14 system. We randomly collect 1.0k samples and use a feedforward neural network with three hidden layers to forecast the load\footnote{More experiment can be found in the appendix.}. We do E2E-AT with \textbf{a).} input feature uncertainties $\bm{\delta}_x$, \textbf{b).} uncertainty of the unpredictable parameters $\bm{\delta}_\phi$, and \textbf{c).} integrated uncertainties of both $(\bm{\delta}_x, \bm{\delta}_\phi)$. We first implement natural (or clean) E2E learning, based on which we warm-start the E2E-ATs. Adam optimizer is used, and `Adversarial training for free' \cite{shafahi2019adversarial} is applied to reuse the gradients for PGD. 

In detail, \textbf{a)}. Since meteorological features have been normalized into $[0,1]$, we attack with a normalized budget $\epsilon_{x} \in \{0.02,0.03\}$. The inner maximization is solved with 7 PGD steps and the step size is dynamically set as $\epsilon_{x} \slash 7 \times 2$. We summarize this setting from \cite{zhang2019theoretically}. We clamp the attacked input into $[0,1]$ whenever it is updated. \textbf{b)}. We consider uncertainties on the susceptance $\bm{b}$ in the \textbf{redispatch} problem. Since each transmission line can have different nominal susceptance, we set the budget $\epsilon_{\phi} \in \{0.05,0.15\}$ as the proportion to the individual nominal value, which is consistent with the common operation range of susceptance \cite{xu2022robust}. \textbf{c)}. We do E2E-AT with $(\epsilon_x, \epsilon_\phi)\in \{(0.02,0.05),(0.03,0.15)\}$ for the integrated uncertainties. The other settings are the same as in (a) and (b).

\subsection{Performance of E2E-AT}

\begin{figure*}[t]
     \centering
     \begin{subfigure}[b]{0.3\textwidth}
         \centering
         \includegraphics[width=\textwidth]{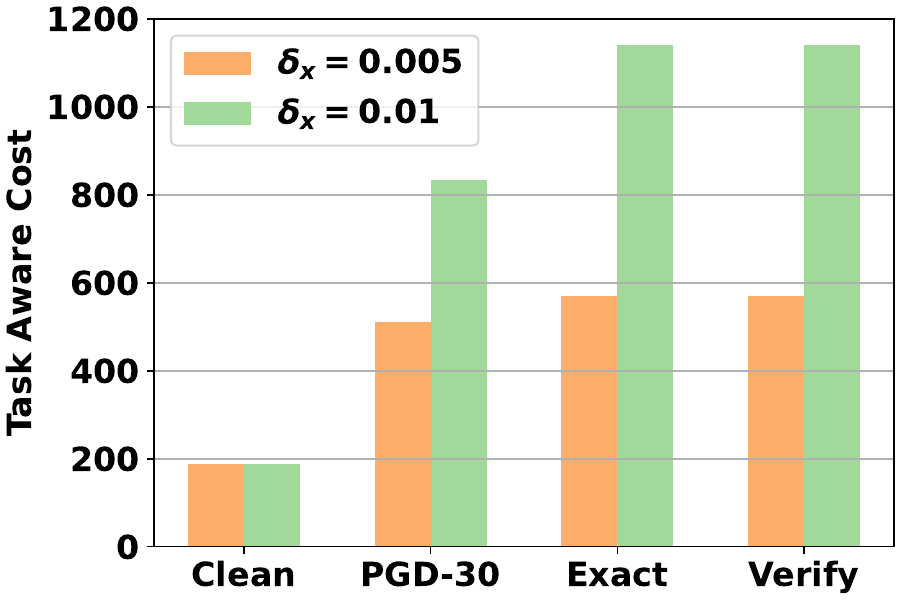}
         \caption{Natural Training}
     \end{subfigure}
     \hfill
     \begin{subfigure}[b]{0.3\textwidth}
         \centering
         \includegraphics[width=\textwidth]{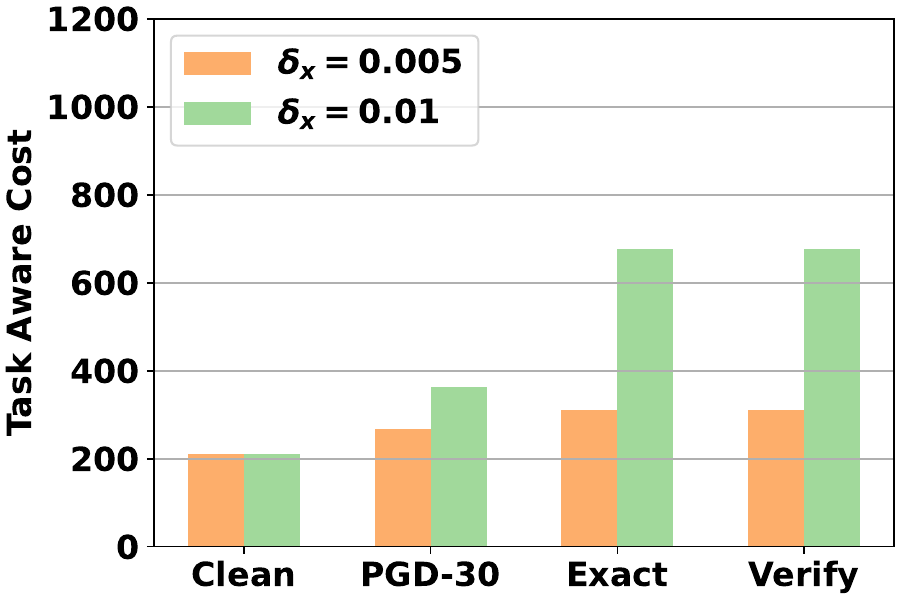}
         \caption{Adversarial Training $\bm{\delta} = 0.02,\alpha=1$}
     \end{subfigure}
     \hfill
     \begin{subfigure}[b]{0.3\textwidth}
         \centering
         \includegraphics[width=\textwidth]{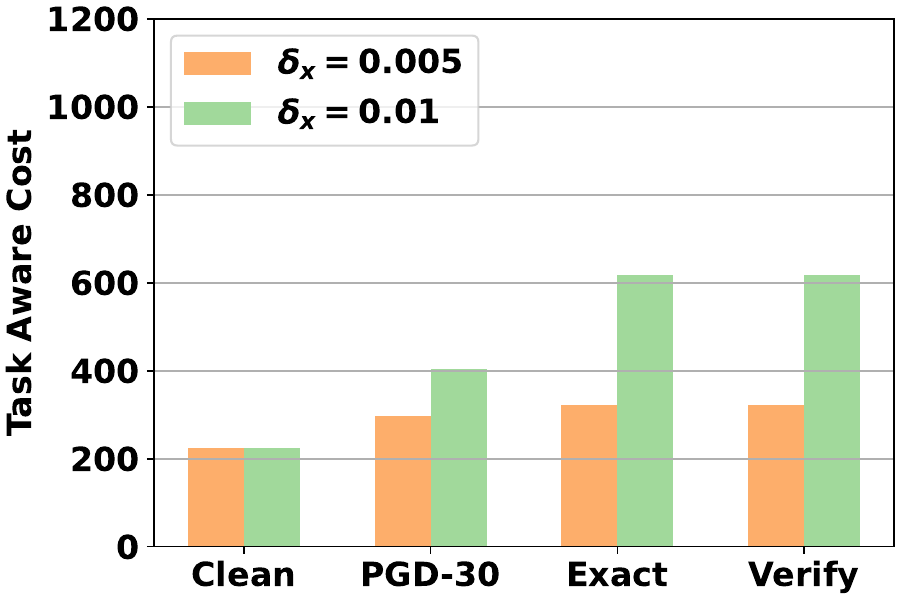}
         \caption{Adversarial Training $\bm{\delta} = 0.03,\alpha=1$}
     \end{subfigure}
        \caption{Input space adversarial attack using the exact MILP formulation. `Clean': cost of the clean sample; `PGD-30': cost of adversarial sample found by 30 PGD steps; `Exact': cost of adversarial sample found by exact MILP; `Verify': cost of the adversarial sample by feeding the MILP solutions into the dispatch and redispatch problems.}
        \label{fig:certify}
\end{figure*}

\begin{table*}[t]
    \centering
    \caption{CO uncertainties with random sampling of $\bm{b}$. The adversarial attack reported in Table \ref{tab:adverarial} is denoted as PGD-7.}
    \scalebox{0.95}{
    \begin{tabular}{c|c|c|c|c|c|c|c}\hline
       \multirow{2}{*}{\textbf{Training Method}}  & \multirow{2}{*}{\textbf{Clean}} & \multicolumn{2}{c|}{$\bm{\epsilon = 0.05}$} & \multicolumn{2}{c|}{\bm{$\epsilon = 0.1$}} & \multicolumn{2}{c}{\bm{$\epsilon = 0.15$}}  \\\cline{3-8}
         & & \textbf{Random} & \textbf{PGD-7} & \textbf{Random} & \textbf{PGD-7} & \textbf{Random} & \textbf{PGD-7} \\\hline\hline
         \textbf{NAT}                                               & 201.6 & 267.2 & 754.6 & 396.6 & 2044.3 & 655.5 & 3468.4     \\\hline
         \textbf{AT-PARA} $\bm{\epsilon_\phi = 0.05,\alpha=1.0}$    & 212.3 & 212.3 & 214.9 & 218.4 & 590.1 & 313.3 & 1931.3\\\hline
         \textbf{AT-PARA} $\bm{\epsilon_\phi = 0.05,\alpha=0.5}$    & 206.7 & 206.7 & 215.6 & 222.6 & 741.2 & 313.1 & 2020.4\\\hline
          \textbf{AT-PARA} $\bm{\epsilon_\phi = 0.15,\alpha=1.0}$    & 214.0 & 214.0 & 214.0 & 214.0 & 221.0 & 222.9 & 453.2\\\hline
         \textbf{AT-PARA} $\bm{\epsilon_\phi = 0.15,\alpha=0.5}$    & 216.5 & 216.5 & 216.5 & 216.5 & 223.1 & 219.4 & 465.6 \\\hline
    \end{tabular}
    }
    \label{tab:random}
\end{table*}

Multi-run adversarial attacks are evaluated in Table \ref{tab:adverarial}. For each sample and attack scenario, we randomly select three starting points within the attack budget and report the \textbf{worst} task-aware cost \eqref{eq:training_objective} to reduce the variance. Specifically, five training algorithms are compared. \textbf{NAT}: natural training with task-aware loss; \textbf{AT-MSE}: adversarial training with MSE loss; \textbf{AT-INPUT}: adversarial training with task-aware loss and input uncertainties; \textbf{AT-PARA}: adversarial training with task-aware loss and unpredictable CO uncertainties; and \textbf{AT-BOTH}: adversarial training with task-aware loss and integrated uncertainties.


We first highlight that uncertainties in COs can significantly increase the cost, e.g. 15 times higher than the clean cost when $\epsilon_\phi = 0.15$ for NAT. The AT-MSE performs better for input attacks, compared to the NAT, but performs poorly on the CO attacks. This is because AT-MSE is only trained with input uncertainties. Second, the performance of E2E-AT is similar to conventional adversarial training for image classification \cite{madry2017towards, zhang2019theoretically}. For instance, training with a larger attack budget can result in better robustness on attacks with a smaller budget but can inevitably increase the clean cost. Meanwhile, the hyperparameter $\alpha$ gives a trade-off between clean and robust accuracy in most cases.

In addition to common findings on adversarial robustness of image tasks, some unique findings of E2E-AT are highlighted. First, AT-INPUT and AT-PARA are more effective in the uncertainty with which they are trained. However, it is observed that E2E-AT based on one source of uncertainty can also improve the robustness of the other. For example, in AT-PARA, the cost of the input attack is even lower than that trained by AT-INPUT, when $\epsilon_x$ is small. Moreover, AT-PARA is more effective in the integrated attack than AT-INPUT. In fact, any input uncertainty eventually feeds into the COs, which becomes the uncertainties of the predictable parameters in COs. 
Finally, AT-BOTH not only improves the robustness of integrated uncertainty, but improves the robustness of the individual's. All of the findings demonstrate that the uncertainties of both sources can be treated together in a unified way.

\subsection{Certified Robustness}

Certification on the robustness of the input space is considered, as susceptance $\bm{b}$ is coupled with the decision variable and Proposition \ref{theorem:qp_kkt} is not applicable. Using Proposition \ref{theorem:qp_kkt} and the mixed integer reformulation of NN \eqref{eq:nn_milp}, the exact input space adversarial attack on sample $(\bm{x},\bm{y})$ can be found by a mixed integer linear program (MILP). Interval bound propagation (IBP) \cite{gowal2018effectiveness} is used to estimate the bounds of layers in NN. As for COs, we set $M=10^5$ by experience. Due to the large computation burden, we randomly sample 15 \textbf{same} samples and solve the robust certification using \texttt{Gurobi}. 

First, the \textit{branch-and-bound} algorithm is applied whose optimality is guaranteed if feasible. Meanwhile, the MILP formulation is verified as the same task-aware cost is achieved when solving the downstream COs parameterized by the optimal attack vector. Second, it can be observed that the exact attack vector causes the worse cost degradation, compared to the PGD-30 attacks. Finally, AT-INPUT can effectively reduce the task-aware cost on the exact input space attack.

\subsection{Parameter Uncertainties in COs}

As shown in \eqref{eq:spo_learning_opt}, the uncertainties of CO can be modeled by stochastic CO. To model stochastic susceptance $\bm{b}$, we randomly alter the susceptance (random attack) for each sample and report the task-aware cost in Table \ref{tab:random}. Under the random attack, the task-aware cost of NAT increases, clearly demonstrating the need for adversarial training. After E2E-AT, task-aware cost under random attacks is significantly reduced and is equal to the corresponding clean cost when the attack budget is small. Although the clean cost increases, it is still lower than the average cost under random attacks. The task-aware cost is also empirically upper bound by the adversarial attack (e.g. PGD-7), which verifies our argument on connecting the stochastic COs with E2E-AT \eqref{eq:robust_formulation} and \eqref{eq:robust_formulation_balance}.

\section{Conclusion}

This paper proposes a unified framework for tackling uncertainties in task-aware E2E learning. We argue that the uncertainties occur at both the input feature of ML and the unpredictable parameter of COs. A robust program is formulated, which is practically solved by adversarial training (E2E-AT). Through theoretical analysis and experiment, we demonstrate that 1). the CO uncertainty can cause significant generalization degradation which has been overlooked before; 2). The optimal adversarial attack on affine-parametric QP can be found by solving the mixed integer (linear) program; and 3). adversarial training can effectively improve the robustness of E2E learning in a unified way.

\section{Acknowledgments}
This research is supported by the Engineering
and Physical Sciences Research Council (EPSRC) under Grant EP/Y025946/1. Wangkun Xu is also supported by the PhD scholarship of the Department of EEE, Imperial College London. Jianhong Wang is fully supported by the UKRI Turing AI World-Leading Researcher Fellowship, EP/W002973/1.

\section{Appendix}

\subsection{Proofs}

\subsubsection{Proof to Proposition \ref{theorem:misleading}}

First, $M(\hat{\bm{z}}_1^\star) \leq M(\hat{\bm{z}}_2^\star)$ can be directly verified since \eqref{eq:form_two} has a tighter constraint on the optimality condition in the lower level problem than it in \eqref{eq:form_one}. For fixed $\bm{\theta}_2^\star$, the optimal decision $\hat{\bm{z}}_r^\star(\bm{\theta}_2^\star)$ is achieved when each subproblem, represented as the lower level problem in \eqref{eq:form_two}, achieves its optimum. Therefore, $\ell(\hat{\bm{z}}_2^{i\star};\bm{y}^i) = \ell(\hat{\bm{z}}_{r}^{i\star}(\bm{\theta}_2^\star);\bm{y}^i)$ and $M(\hat{\bm{z}}_2^\star) = M(\hat{\bm{z}}_r^\star(\bm{\theta}_2^\star))$. It also shows that $(\bm{\theta}_2^\star,\bm{z}_r^\star(\bm{\theta}_2^\star))$ is the minimizer of \eqref{eq:form_two}. Note that multiple global minimizers are also satisfied. Since $(\bm{\theta}_1^\star,\bm{z}_r^\star(\bm{\theta}_1^\star))$ is also feasible with \eqref{eq:form_two}, it gives $M(\hat{\bm{z}}_r^\star(\bm{\theta}_2^\star)) \leq M(\hat{\bm{z}}_r^\star(\bm{\theta}_1^\star))$, which finalizes the proof.

\subsubsection{Proof to Proposition \ref{theorem:qp_kkt}}

We give the proof by first introducing the complementary linearization \cite{fortuny1981representation,kazempour2010strategic}:
\begin{proposition}\label{theorem:linearization}
    The complementary condition $a \geq 0, b \geq 0, a\cdot b = 0$ can be replaced by
    \begin{equation*}
        a \geq 0, \; b \geq 0, \; a \leq \psi M, \; b \leq(1-\psi) M, \; \psi \in\{0,1\}
    \end{equation*}
where $M$ is a large enough constant.
\end{proposition}

For the optimal primal-dual pair $(\bm{z}_{i+1}, \bm{\lambda}_{i+1}, \bm{\nu}_{i+1})$, the Karush–Kuhn–Tucker (KKT) condition \cite{boyd2004convex} gives that 
\begin{subequations}
    \begin{equation}
        \bm{Q}\bm{z}_{i+1} + \bm{q} + \bm{A}^T\bm{\lambda}_{i+1} + \bm{C}^T\bm{\nu}_{i+1} = 0 
    \end{equation}
    \begin{equation}
        \bm{C}\bm{z}_{i+1} + \bm{H}\bm{z}_i - \bm{d} = 0
    \end{equation}
    \begin{equation}\label{eq:kkt_complementary}
        \operatorname{diag}(\bm{\lambda}_{i+1})(\bm{A}\bm{z}_{i+1} + \bm{G}\bm{z}_i + \bm{b}) = 0
    \end{equation}
    \begin{equation}\label{eq:kkt_inequality}
        \bm{A}\bm{z}_{i+1} + \bm{G}\bm{z}_i - \bm{b} \leq 0
    \end{equation}
    \begin{equation}\label{eq:kkt_lambda}
        \bm{\lambda}_{i+1} \geq 0
    \end{equation}    
\end{subequations}

Each of the lower-level problems can be equivalently written in this form. Since the complementary condition states that at least one of the inequality constraints or the dual variable equals zero, \eqref{eq:kkt_complementary}, \eqref{eq:kkt_inequality}, and \eqref{eq:kkt_lambda} can be rewritten by Proposition \ref{theorem:linearization}, which finalizes the proof.

\subsection{Network Constrained Economic Dispatch}

The stage-one problem is a generator dispatch problem, which can be modelled as:
\begin{equation}\label{eq:power_stage_one}
    \begin{array}{rl}
        (\bm{P}_g^\star, \bm{\vartheta}^\star, \bm{s}^\star) & = \arg\min_{\bm{P}_g, \bm{\vartheta}, \bm{s}} \bm{c}_g^T\bm{P}_g + c_{ls} \bm{1}^T \bm{s} \\
        \text{s.t. } & \underline{\bm{P}}_{g} \leq \bm{P}_g \leq \bar{\bm{P}}_g\\
                    & \bm{A}^T\operatorname{diag}(\bm{b})\bm{A}\bm{\vartheta} = \bm{C}_g\bm{P}_g - \bm{C}_l(\hat{\bm{y}} - \bm{s}) \\
                    & \underline{\bm{P}}_f \leq \operatorname{diag}(\bm{b})\bm{A}\bm{\vartheta} \leq \bar{\bm{P}}_f \\
                    & \bm{s} \geq 0 \\
                    & \bm{\vartheta}_{\text{ref}} = 0
    \end{array}
\end{equation}

In \eqref{eq:power_stage_one}, $\bm{P}_g$ and $\bm{\vartheta}$ are the vector of generator dispatch and voltage angle in each bus. A linear cost function is considered. All the equality and inequality signs in the constraints are element-wise. The first constraint represents the upper and lower bounds for each generator. The second constraint represents the thermal limit on the power flow in each transmission line where $\bm{b}$ is the line susceptance, $\bm{A}$ is the incidence matrix of the power system. The third constraint represents the requirement for power balance on each bus, where $\bm{C}_g$ and $\bm{C}_l$ are the incidence matrices of the generator and the load, respectively. $\hat{\bm{y}}$ represents the forecast load which is parameterized by the neural network model, for example, $\hat{\bm{y}} = f(\bm{x};\bm{\theta})$. We also add a slack variable $\bm{s}\geq\bm{0}$ as a compensation for infeasibility with large cost coefficient $c_{ls}$. The fifth constraint indicates that the system is referenced to the slack bus with constant phase angle. 

When the generator has been dispatched as $\bm{P}_g^\star$, we consider a realization on the actual load $\bm{y}$ by solving the second stage problem (also known as redispatch problem):
\begin{equation}
    \begin{array}{rl}
        (\bm{P}_{ls}^\star, \bm{P}_{gs}^\star, \bm{\vartheta}^\star) &= \arg\min_{\bm{P}_{ls}, \bm{P}_{gs}, \bm{\vartheta}} \bm{c}_{ls}^T\bm{P}_{ls} + \bm{c}_{gs}^T\bm{P}_{gs} \\
        \text{s.t. }
                    & \bm{A}^T\operatorname{diag}(\bm{b})\bm{A}\bm{\vartheta} = \bm{C}_g(\bm{P}_g^\star - \bm{P}_{gs}) \\
                    & \qquad \qquad \qquad \qquad -\bm{C}_l({\bm{y}} - \bm{P}_{ls}) \\
                    & \underline{\bm{P}}_f \leq \operatorname{diag}(\bm{b})\bm{A}\bm{\vartheta} \leq \bar{\bm{P}}_f \\
                    & \bm{P}_{ls} \geq 0, \bm{P}_{gs} \geq 0 \\
                    & \bm{\vartheta}_{\text{ref}} = 0
    \end{array}
\end{equation}

The objective of stage two is to balance the load and to solve any violation of physical constraints of power grid using load shedding $\bm{P}_{ls}$ if the generator dispatch is lower than the actual load and energy storage $\bm{P}_{gs}$ if the generator dispatch is higher than the actual load. Since load shedding (similar to blackout) is more critical and should be avoided as much as possible, it is assigned by a larger penalty, i.e. $\bm{c}_{ls} \gg \bm{c}_{gs}$.

We highlight that our formulation on power system operation is more realistic than \cite{donti2017task}, in which the behavior of different loads and network constraints are ignored. This results in more complex COs. In detail, the two COs have more than 60 decision variables and 150 constraints in total, which needs to be exactly solved for every forward pass.

\subsection{Certifying the Load Forecasting E2E Learning}

Using Proposition \ref{theorem:qp_kkt} and the mixed integer reformulation of NN \eqref{eq:nn_milp}, an exact adversarial attack on input $(\bm{x},\bm{y})$ can be found by MILP. It is known that the value of the `big M' used for NN linearization (e.g. the lower and upper bounds of the output of each NN layer \eqref{eq:nn_milp}, the upper bounds of the active inequality dual variables, and the lower bounds of the inactive inequality constraints of the COs) is essential to the performance of MILP solution. Therefore, interval bound propagation (IBP) \cite{gowal2018effectiveness} is used to estimate the layers' bounds in NN. As for COs, we set $M=10^5$ by experience.

\subsubsection{Formulation}

By using Proposition \ref{theorem:qp_kkt} and the mixed integer reformulation of NN \eqref{eq:nn_milp}, the exact input space adversarial attack on sample $(\bm{x},\bm{y})$ can be found by an MILP:
\begin{equation}\label{eq:certify_load}
    \begin{array}{rl}
        \bm{\delta}^\star = \operatorname{max}_{\bm{\delta}} & \bm{c}_g^T\bm{P}_{g} + \bm{c}_{ls}^T\bm{P}_{ls} + \bm{c}_{gs}^T\bm{P}_{gs} \\
        \text{subject to} 
        & \bm{P}_{ls}, \bm{P}_{gs} \in \mathcal{C}_{\text{Lin-KKT}}^{\text{Reispatch}}(\bm{P}_{ls}, \bm{P}_{gs};\bm{P}_g)\\
        & \bm{P}_g \in \mathcal{C}_{\text{Lin-KKT}}^{\text{Dispatch}}(\bm{P}_g;\hat{\bm{y}}) \\
        & \hat{\bm{y}} \in \mathcal{C}_{\text{nn}}(\bm{x}+\bm{\delta};\bm{\theta})
    \end{array}
\end{equation}
where $\mathcal{C}_{\text{nn}}(\cdot)$ is the mixed integer linear representation of the trained neural network by \eqref{eq:nn_milp}, $\mathcal{C}_{\text{Lin-KKT}}^{\text{Dispatch}}(\cdot)$ and $\mathcal{C}_{\text{Lin-KKT}}^{\text{Reispatch}}(\cdot)$ are the linearized KKT conditions of the lower level dispatch and redispatch problems, respectively by Proposition \ref{theorem:qp_kkt}. Note that \eqref{eq:certify_load} is a rather simplified representation that omits the detailed formulation of the constraints, the integer variables in $\mathcal{C}_{\text{nn}}(\cdot)$, $\mathcal{C}_{\text{Lin-KKT}}^{\text{Dispatch}}$, and $\mathcal{C}_{\text{Lin-KKT}}^{\text{Reispatch}}$, dual variblaes, as well as the associated lower and upper bounds (big-M) of the integers. Nonetheless, \eqref{eq:certify_load} is formulated as MILP, which can be solved by solvers like Gurobi.

\subsubsection{Interval Bound Propagation (IBP)}

Based on \eqref{eq:nn_milp}, the lower and upper bounds of each layer in NN can be estimated linearly as:
\begin{equation}
    \begin{split}
    \hat{\bm{l}}_i & = \max{\{\bm{0}, \bm{l}_i\}} \\
        \hat{\bm{u}}_i & = \max{\{\bm{0}, \bm{u}_i\}} \\
        \bm{l}_{i+1} & = \max{\{\bm{0}, \bm{W}_i\}}\cdot \hat{\bm{l}}_i + \min{\{\bm{0},\bm{W}_i\}}\cdot \hat{\bm{u}}_i + \bm{b}_i\\
        \bm{u}_{i+1} & = \min{\{\bm{0}, \bm{W}_i\}}\cdot \hat{\bm{l}}_i + \max{\{\bm{0},\bm{W}_i\}}\cdot \hat{\bm{u}}_i + \bm{b}_i
    \end{split}
\end{equation}
for $i=1,\cdots,d-1$. The initial bound is determined by the attack budget $\epsilon$, that is, $\hat{l}_1=\bm{x}-\epsilon\cdot\bm{1}$ and $\hat{u}_1=\bm{x}+\epsilon\cdot\bm{1}$.

\subsection{Further Analysis on the Relationship between Input and CO Uncertainties}

Based on the experiment results in Table \ref{tab:adverarial} and Table \ref{tab:adverarial_test}, we speculate that the robustness of one uncertainty can improve the robustness of the other, that is, the two uncertainties are not contradictory in E2E-AT. We thereby give a more detailed discussion.

To start, the two sources of uncertainties, e.g., input and CO uncertainties, are handled by training a \textbf{same} robust NN, which is a parametric model to forecast the predictable parameter of CO. According to the sensitivity analysis \cite{boyd2004convex}, a robust E2E model under uncertain COs implies that the robust NN can forecast a parameter such that the activations of constraints will not be significantly changed. From the structure of E2E learning, the input uncertainty is amplified by the NN \cite{gowal2018effectiveness}, which becomes an additional uncertain parameter of the COs. Similarly, an input-robust E2E model also requires that the NN forecast does not trigger significant constraint violations. Therefore, we speculate that the E2E-AT training objectives under input and CO uncertainties are not contradictory, as they are both reflected and controlled by the behavior of the COs.

However, the effectiveness of AT-INPUT on CO uncertainties (or AT-PARA on input uncertainties) depends on the individual attack budget, which will be discussed later by the extra experiment.

\subsection{Details on experiment settings}

\subsubsection{Reproducibility}

Our experiment is reproducible and open source on GitHub\footnote{\url{https://github.com/xuwkk/E2E-AT}.}.

\subsubsection{Data Source}

The IEEE bus-14 system is modified from \texttt{PyPower}\footnote{\url{https://github.com/rwl/PYPOWER/blob/master/pypower/case14.py}.}.

The meteorological features in the Texas Backbone Power System \cite{lu2023synthetic} include temperature (k), longwave radiation (w / m2), shortwave radiation (w / m2), zonal wind speed (m / s), meridional wind speed (m / s) and wind speed (m / s) which are normalized into [0,1]. The calendar feature includes the cosine and sin of the weekday in a week and hour in a day according to their individual period. We pack the meteorological features of 14 buses as well as the 4 calendric features. Therefore, a single datum is $(\bm{x}^i, \bm{y}^i)\in\mathbf{R}^{4+6*14}\times \mathbf{R}^{14}$. We map the dataset to the scale that is suitable for the bus-14 system. In detail, we start at small ground-truth load profile and gradually increase to just have the feasible solution of the dispatch and redispatch problems. 

\subsubsection{Packages}

During inference, we formulate the dispatch and redispatch problem by \texttt{Cvxpy} \cite{diamond2016cvxpy}, which are solved by calling \texttt{Gurobi}\footnote{\url{https://www.gurobi.com}.}. When calculating the gradient, we use \texttt{PyTorch} automatic differentiation package and \texttt{CvxpyLayers} to implement fast batched forward and backward passes\cite{agrawal2019differentiable}.

\subsubsection{Network structure}

The forecast neural network has three hidden layers with output sizes of 200, 200, and 100. ReLU activations are added between layers. We also add a ReLU activation before the convex layer so that the forecast load is guaranteed to be positive and the adversarial attack cannot result in negative forecast load as well. The ReLU layer is also added when evaluating the certified approach.

\subsubsection{Training Settings}

For all experiments, we set batch size as 32 and use Adam optimizer \cite{kingma2014adam}. We train the NN with MSE loss for 250 epochs. The learning rate is $10^{-3}$ with cosine annealing. We store the NN states at 200 epochs and warm-start natural E2E learning for 50 epochs with learning rate $10^{-5}$. For E2E-AT, we warm-start training from the state trained by natural E2E learning for 100 epochs with learning rate $10^{-5}$. As we set the PGD step to be 7, it is equivalent to 14 epochs under adversarial training for free setting.

For the input space attack, we only attack the meteorological features in the input as the calendric features are discrete and can be easily verified by the operator. For uncertainties in the unpredictable parameter in COs, the susceptance of the transmission line is attacked. This is a realistic setting, as the susceptance of the transmission lines may vary due to temperature changes or can be intentionally altered by the system operator via electronic devices \cite{xu2022robust}. However, such changes cannot be detected when forecasting the load and during the dispatch stage.

In addition, we noticed that the objective of E2E adversarial training is nonsmooth. For example, the training objective can be significantly increased when certain inactive CO constraints become active or vice versa. Therefore, we use gradient clip to restrict the 1-norm of the gradient to be maximum 2.0 when updating the network. We found that this setting can result in more stable training.

\subsection{Extra Experiment Results}

\subsubsection{Robust Performance on the Test Dataset}

In the main content, we train on 1.0k random samples in the Texas backbone power system and report the clean and robust task-aware cost. Here we train on the entire dataset which contains 8760 samples (i.e., one hour resolution of year 2019). We do random train-test split with proportion 8:2. The training epoch is increased to 200 (or 28 epochs using `adversarial training for free') and the batch size is increased to 64. The remaining training settings are the same as before. We then attack the trained model and report the performance on the \textbf{test} dataset. 

To observe some new results, we increase the attack budget of the input space uncertainty as $\epsilon_x\in[0.015, 0.04, 0.06]$ to have a task-aware cost comparable to the parameter attack and set the parameter attack budget as $\epsilon_\phi\in[0.05,0.10,0.15]$. Furthermore, We report the worst task-aware cost within three multi-runs. Note that we select the worst attack vector for \textbf{each} sample in the minibatch. 

The extra experiment results are shown in Table \ref{tab:adverarial_test}. In general terms, the extra experiment in the large data set illustrates similar results to those in Table \ref{tab:adverarial}. And the main difference is caused by the increase in the input attack budget $\epsilon_x$.

\begin{table*}
    \centering
    \footnotesize
    \caption{Performances of the E2E-AT. The model is trained on a larger dataset and evaluated on the test dataset.}
    \scalebox{0.95}{
    \begin{tabular}{c|c|c|c|c|c|c|c|c|c|c|c|c}\hline
       \multicolumn{3}{c|}{\textbf{Training Method}} & \textbf{Clean} & \multicolumn{3}{c|}{\textbf{Input Attack}, $\epsilon_x$} & \multicolumn{3}{c|}{\textbf{CO Attack}, $\epsilon_\phi$} & \multicolumn{3}{c}{\textbf{Integrated Attack}, ($\epsilon_x, \epsilon_\phi$)}   \\\hline\hline
       $\epsilon_x$ & $\epsilon_\phi$ & $\alpha$ & N/A & 0.015 & 0.04 & 0.06 & 0.05 & 0.10 & 0.15 & (0.015,0.05) & (0.04,0.10) & (0.06,0.15) \\\hline\hline

       \multicolumn{13}{c}{\textbf{NAT: Natural Training with Task Loss}} \\\hline\hline
       0 & 0 & 0 & 220.3 & 943.8 & 2366.2 & 3528.0 & 852.8 & 2254.5 & 3679.3 & 1526.4 & 4287.7 & 6477.6 \\\hline\hline
       
        \multicolumn{13}{c}{\textbf{AT-MSE: Adversarial Training with MSE Loss}} \\\hline\hline
       0.015 & N/A & N/A & 356.9 & 836.0 & 1792.3 & 2552.5 & 1294.3 & 2705.0 & 4069.1 & 1727.0 & 3864.4 & 5844.9  \\\hline
       0.04 & N/A & N/A & 462.3 & 745.3 & 1300.0 & 1820.6 & 1292.0 & 2647.1  & 4024.5 & 1558.3 & 3414.7 & 5273.2 \\\hline
       0.06 & N/A & N/A & 646.4 & 846.5 & 1229.5 & 1586.8 & 1352.7 & 2652.2 & 4022.7 & 1540.6 & 3169.1 &  4967.4\\\hline\hline
       
       \multicolumn{13}{c}{\textbf{AT-INPUT: Adversarial Training with Task Loss on the Input Uncertainty}} \\\hline\hline
       \multirow{2}{*}{0.015} & \multirow{2}{*}{0} & 1.0 & 302.1 & 430.3 & 692.3 & 981.3 & 369.2 & 884.5 & 1939.9 & 488.3 & 1374.5 &  2981.5 \\\cline{3-13}
                                                 &  & 0.5 & 240.9 & 544.4 & 1322.8 & 2009.5 & 756.5 & 2038.0 & 3455.0 & 1074.3 & 3361.1 & 5566.8 \\\hline
       \multirow{2}{*}{0.04} & \multirow{2}{*}{0} & 1.0 & 429.1 & 539.4 & 755.7 & 950.9 & 449.5 & 517.3 & 656.0 & 552.1 & 802.2 & 1050.6 \\\cline{3-13}
                                                 &  & 0.5 & 280.9 & 509.7 & 972.0 & 1403.5 & 739.3  & 1875.1 & 3259.2 & 940.9 & 2687.7 & 4647.4 \\\hline
       \multirow{2}{*}{0.06} & \multirow{2}{*}{0} & 1.0 & 436.4 & 528.2 & 706.5 & 859.0 & 449.6 & 490.7 & 584.6 & 538.1 & 729.9 & 940.9 \\\cline{3-13}
                                                 &  & 0.5 & 307.8  & 432.1 & 659.9 & 863.5 & 553.0  & 1350.5 & 2500.5 & 660.0 & 1788.9 & 3268.0   \\\hline\hline
                                                 
       \multicolumn{13}{c}{\textbf{AT-PARA: Adversarial Training with Task Loss on the CO Uncertainty}} \\\hline\hline
       \multirow{2}{*}{0} & \multirow{2}{*}{0.05} & 1.0 & 223.0 & 1003.9 & 2532.6 & 3717.9 & 231.1 & 967.3 & 2526.4 & 1014.9 & 2908.2 & 5028.1 \\\cline{3-13}
                                             &  & 0.5 & 224.8 & 1019.5 & 2538.8 & 3717.7 & 239.0  & 1096.5 & 2677.8 & 1028.5 & 3000.9 & 5222.0\\\hline
       \multirow{2}{*}{0} & \multirow{2}{*}{0.10} & 1.0 & 227.3 & 934.2 & 2456.0 & 3653.9 & 227.3 & 262.7 & 956.7 & 949.4 & 2483.2 & 3912.2 \\\cline{3-13}
                                             &  & 0.5 & 226.9 & 927.3 & 2325.5 & 3392.5 & 227.1 & 269.6 & 1072.2 & 928.5 & 2326.8 & 3760.9 \\\hline
       \multirow{2}{*}{0} & \multirow{2}{*}{0.15} & 1.0 & 244.8 & 1042.6 & 2552.3 & 3737.4 & 244.8 & 245.5 & 269.1 & 1039.3 & 2545.7 & 3847.1 \\\cline{3-13}
                                             &  & 0.5 & 237.1 & 974.3 & 2470.4 & 3646.2 & 237.1 & 237.7 & 286.6 & 978.2 & 2455.3 & 3733.9 \\\hline\hline

       \multicolumn{13}{c}{\textbf{AT-BOTH: Adversarial Training with Task Loss on the Integrated Uncertainties}} \\\hline\hline
       \multirow{2}{*}{0.015} & \multirow{2}{*}{0.05} & 1.0 & 305.9 & 447.8 & 734.3 & 1063.7 & 308.7 & 389.2 & 643.5 & 450.2 & 816.0 & 1296.1 \\\cline{3-13}
                                                 &  & 0.5 & 267.2 & 743.7 & 1734.4 & 2593.4 & 404.5 & 1671.6 & 3150.6 & 806.5 & 2865.1 & 5163.7\\\hline
       \multirow{2}{*}{0.04} & \multirow{2}{*}{0.10} & 1.0 & 385.4 & 462.6 & 600.0 & 731.7 & 385.8 & 394.0 & 438.8 & 461.4 & 608.8 & 782.4 \\\cline{3-13}
                                                 &  & 0.5 & 293.3 & 516.2 & 961.2 & 1376.6 & 296.4 & 456.6 & 1069.9 & 516.2 & 1061.1 & 1849.6 \\\hline\hline
    \end{tabular}
    }
    \label{tab:adverarial_test}
\end{table*}

We summarize some of the findings below.
\begin{enumerate}
    \item In all E2E-AT, the clean accuracy decreases as the attack budget increases, which is consistent to the conventional adversarial training. Moreover, model trained under larger attack budget can also improve robustness under smaller attack budget.
    \item Hyperparameter $\alpha$ can balance clean accuracy and robust accuracy, which is consistent with conventional adversarial training.
    \item The AT-MSE can impact the robust accuracy under input attack (but still much higher than E2E-AT), but cannot improve the robustness of CO, which actually becomes worse compared to NAT. This is because the AT-MSE only captures the input uncertainties and fits to the MSE loss. Therefore, it cannot be generalized to the uncertainty of CO.
    \item Both input and CO uncertainties must be considered when designing the E2E learning task. E2E-AT is an effective approach to improve the model robustness. And the unified training (AT-BOTH) can result in the best task-aware costs in most of the cases.
    \item In the new experiment, it can be observed that training under input space adversaries can improve not only the input robustness but the CO robustness. We actually have opposite results of the experiment in Table \ref{tab:adverarial} in which the robustness of CO can improve the robustness of the input. We argue that the main reason is caused by the different influence of the input and CO uncertainties under different attack budgets. This explains why AT-BOTH can significantly improve the robustness of CO but only has a limited improvement on the robustness of input uncertainty. We will discuss this point in the next section.
    
\end{enumerate}

\subsubsection{Gradient Analysis of E2E-AT}

From Table \ref{tab:adverarial_test}, we compare the uncertainty of the input and unpredictable parameters in CO by designing the attack budget $\epsilon_x$ and $\epsilon_\phi$ to have a similar task-aware cost in the clean E2E model. We conclude that the robustness of CO is easier to improve than the robustness of input, and improving the robustness of input can improve the robustness of CO. To verify this point, we calculate the average $\ell_1$ norm of the gradient of the NN when input and unpredictable parameter attacks are \textbf{separately} generated. Both the NN without E2E-AT (warm started by E2E learning) and the NN updated by AT-BOTH (with $\epsilon_x = 0.04$ and $\epsilon_\phi = 0.10$) are evaluated.

First, as shown in Table \ref{tab:adver_grad}, the gradient of E2E-AT is very large and requires a gradient clip during training. Second, although the initial task-aware costs are similar for input and parameter adversaries (2366 vs 2354), the initial gradient under input adversaries is 10 times higher than it of parameter adversaries. Even after AT-BOTH, the difference is still more than 10 times. This implies that the impact of the CO uncertainties on the task-aware cost is much less than the input uncertainties under the current attack budget. 

\begin{table}
    \centering
    \footnotesize
    \caption{Un-clipped average gradients for E2E-AT in $\ell_1$-norm.}
    \scalebox{0.95}{
    \begin{tabular}{c|c|c}
         &  \textbf{Before E2E-AT} & \textbf{After E2E-AT} \\\hline\hline
       \textbf{Input Attack}  & $4.5\times10^5$ & $8.4\times10^4$ \\\hline
       \textbf{CO Attack}  & $5.8\times10^4$  & $5.7\times10^3$ \\\hline
    \end{tabular}
    }
    \label{tab:adver_grad}
\end{table}

These findings do not violate our initial goal of unifying uncertainties in E2E learning. Actually, we can show that the improvements in the two uncertainties are not contradictory. Otherwise, improving the uncertainties of one cannot improve the uncertainty of the other. To verify the idea, we plot the cosine similarities of the gradients of NN for all mini-batches of size 16 on the entire training dataset.

\begin{figure}
     \centering
     \begin{subfigure}[b]{0.6\linewidth}
         \centering
         \includegraphics[width=\textwidth]{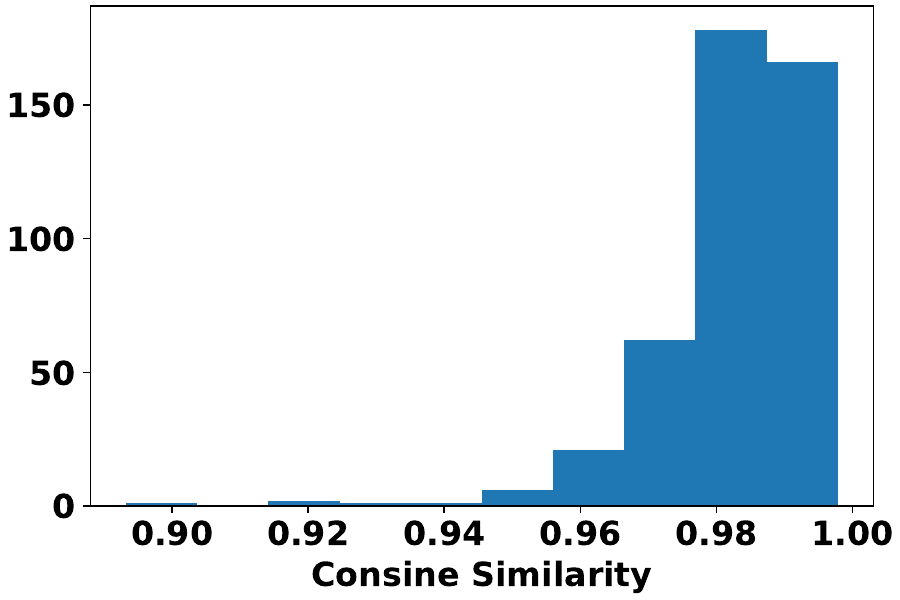}
         \caption{Before E2E-AT (warm start from E2E learning)}
     \end{subfigure}
     \hfill
     \begin{subfigure}[b]{0.6\linewidth}
         \centering
         \includegraphics[width=\linewidth]{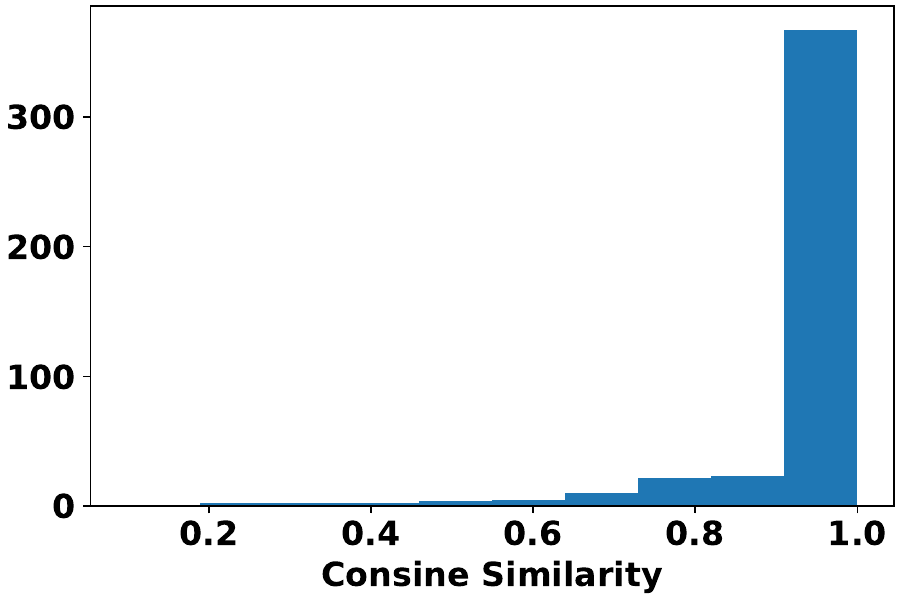}
         \caption{After E2E-AT}
     \end{subfigure}
        \caption{Cosine similarities of the gradients of NN under the input space adversarial attack and unpredictable parameter attack in CO.}
        \label{fig:gradient_similarity}
\end{figure}

As shown in Fig.\ref{fig:gradient_similarity}, the cosine similarities are close to one, meaning that training on one of the uncertainties can also improve the robustness of the other, which is the same as observed by the experiment results.

From the experiment results in Table \ref{tab:adverarial} and \ref{tab:adverarial_test}, as well as the discussions before, we conjecture that, assuming the training objectives on improving the robustness of different uncertainties sources are not contradictory:
\begin{enumerate}
    \item The gradients of NN under the adversaries/uncertainties from different sources can be used to model their impacts on E2E-AT. It can also give a hint on setting the individual attack budget. And to have more balanced adversarial training behaviors among different uncertainty sources, the attack budgets might be set according to the initial NN gradients other than the task-aware costs. 
    \item The dominating uncertainty, for example, the one with a higher NN gradient, can dominate the training process and be also effective on the remaining uncertainties. On the contrary, uncertainty with a smaller impact cannot significantly improve the robustness of the others.
\end{enumerate}

\section{Future Work}

In addition to conjectures, a thorough investigation of the relationship between \textbf{ multiple sources of uncertainty} could be an interesting future work, since conventional adversarial training usually has a single source of uncertainty. Theoretical analysis is needed and \textit{Multi-task learning} \cite{yu2020gradient}, which simultaneously trains for various objectives, can be borrowed to handle multi-uncertainties. In addition, solving the exact attack vector is an MILP problem which cannot be used for adversarial training due to its high complexity. Developing certified and tractable E2E-AT is also important for security purposes.

\bibliography{aaai24}

\end{document}